\documentclass[10pt, a4paper]{article}

\usepackage[final]{lrec2026} %

\usepackage{microtype}
\usepackage{hyperref}
\usepackage{url}
\usepackage{booktabs}
\usepackage{amsmath}
\usepackage{amssymb}
\usepackage{graphicx}
\usepackage{makecell}
\usepackage{cleveref}
\usepackage{tabularx}
\usepackage[shortlabels]{enumitem}
\usepackage{placeins} %

\usepackage[acronym]{glossaries}
\glsdisablehyper

\newacronym{LM}{LM}{Language Model}
\newacronym{LLM}{LLM}{Large Language Model}
\newacronym{QA}{QA}{Question Answering}
\newacronym{BPE}{BPE}{Byte-Pair Encoding}
\newacronym{BBPE}{BBPE}{Byte-Level BPE}
\newacronym{SP}{SP}{SentencePiece}
\newacronym{HF}{HF}{Hugging Face}
\newacronym{NLP}{NLP}{Natural Language Processing}

\title{Diagnosing Translated Benchmarks: An Automated Quality Assurance Study of the EU20 Benchmark Suite}

\name{
    Klaudia Thellmann\textsuperscript{1}\(^\dagger\),
    Bernhard Stadler\textsuperscript{1,2}\(^\dagger\),
    Michael Färber\textsuperscript{1}
}

\address{
    \textsuperscript{1}TU Dresden and ScaDS.AI, \textsuperscript{2}InfAI e.V.\\
    \{klaudia-doris.thellmann, bernhard.stadler, michael.faerber\}@tu-dresden.de\\
    \(^\dagger\) Main authors
}

\abstract{
Machine-translated benchmark datasets reduce costs and offer scale, but noise, loss of structure, and uneven quality weaken confidence.
What matters is not merely whether we can translate, but also whether we can measure and verify translation reliability at scale.
We study translation quality in the EU20 benchmark suite, which comprises five established benchmarks translated into 20 languages, via a three-step automated quality assurance approach:
(i) a structural corpus audit with targeted fixes;
(ii) quality profiling using a neural metric (COMET, reference-free and reference-based) with translation service comparisons (DeepL / ChatGPT / Google);
and (iii) an LLM-based span-level translation error landscape.
Trends are consistent: datasets with lower COMET scores exhibit a higher share of accuracy/mistranslation errors at span level (notably HellaSwag; ARC is comparatively clean).
Reference-based COMET on MMLU against human-edited samples points in the same direction.
We release cleaned/corrected versions of the EU20 datasets, and code for reproducibility.
In sum, automated quality assurance offers practical, scalable indicators that help prioritize review -- complementing, not replacing, human gold standards.%
}

\begin{document}

\maketitleabstract

\section{Introduction}
\label{sec:intro}

Large Language Models (LLMs) have transformed NLP, yet rigorous multilingual evaluation remains challenging beyond high-resource settings.
Across Europe, language- and region-specific suites have emerged, covering e.g.~Scandinavian languages~\citep{nielsen_2023}, Norwegian~\cite{samuel_norbench_2023}, German~\citep{pfister_2024}, Italian~\citep{Evalita-LLM_2025}, and Iberian languages~\citep{baucells_2025}, Czech~\citep{fajcik_2025}, Polish~\citep{LLMzSz_2025}, Greek~\citep{peng_plutus_2025}, and French~\citep{faysse_2025}.
These native resources improve quality and task relevance, but heterogeneity in scope, construction protocol, and task mix limits parallelism and cross-language comparability at scale~\cite{ott2022mapping,srivastava_2024,yang-etal-2019-paws}

Translating existing benchmarks automatically is a pragmatic alternative that scales, but concerns about translation noise, loss of structure, and uneven quality limit trust in such evaluations~\citep{plaza_2024,meng2022generating,nllbteam2022language}.
As a result, the question is not merely whether we can translate benchmarks, but whether machine-translated benchmarks meet quantifiable reliability and diagnostic criteria to guide LLM development and cross-language comparison of LLMs at scale.

We ground our study in EU20~\citep{thellmann_2024}, which translates five established English benchmarks into 20 European languages using DeepL.
While the EU20 benchmark suite offers scale and coverage, comprehensive quality assurance, whether human-based or automated, was not the primary focus of the initial release.
We take a first step toward scalable validation for EU20 by combining structural diagnostics with two complementary, automated translation quality estimation (TQE) methods (i) neural quality estimation based on COMET scores~\citep{rei-etal-2020-comet,rei-etal-2023-scaling,guerreiro-etal-2024-xcomet} and (ii) an LLM-as-a-judge procedure~\citep{kocmi-federmann-2023-large}.
Automated TQE does not replace expert human review, but helps prioritize where to look first (e.g., accuracy vs.~fluency issues) and provides indicators of translation quality under budget constraints.

In this paper, we operationalize translation quality validation on EU20 along three dimensions:
\begin{enumerate}
    \item \textbf{Structural integrity}: a corpus-level audit of field completeness, split/subset consistency, and cross-language coverage (\Cref{sec:struct_audit_repair}).
    \item \textbf{Item-level quality profiling}: a quality landscape, by task and language, for translated benchmark entries using reference-free and reference-based xCOMET-XXL \citep{guerreiro-etal-2024-xcomet}, including paired comparisons across distinct translation services -- EU20/DeepL\footnote{\url{https://www.deepl.com}}, Okapi/ChatGPT\footnote{\url{https://openai.com/research/gpt-4}}~\cite{lai-etal-2023-okapi}, and Global-MMLU/Google Translate\footnote{\url{https://translate.google.com}}~\citep{singh_2025_global} (\Cref{sec:eu20-analysis}).
    \item \textbf{Span-level diagnostic validity}: an interpretable error landscape from an LLM-as-a-judge TQE setup, quantifying error categories (Ac\-cu\-ra\-cy/Mis\-trans\-la\-tion, Flu\-en\-cy, O\-th\-er) and severities across languages and tasks, and testing convergence with xCOMET-XXL (\Cref{subsec:eu20_gemba}).
\end{enumerate}

To make this operationalization actionable and reproducible, we provide the following artifacts:
\begin{itemize}
    \item A cleaned and corrected version of EU20 with documented fixes from our structural audit\footnote{\url{https://hf.co/eu20-cleaned/datasets}} (\Cref{sec:struct_audit_repair}).
    \item Code for the structure-preserving cleaning and correction used in the audit, enabling reproducible maintenance of the EU20 benchmark suite\footnote{\url{https://github.com/eu20-cleaned/lang-integrity}}.
    \item Our LLM-as-a-judge TQE setup (prompts, few-shot exemplars, and scripts) for span-level error annotation\footnote{\url{https://github.com/eu20-cleaned/translation-quality-analysis}}.
\end{itemize}

\section{Related Work}
\label{sec:rel_work}
Prior work on European LLM evaluation spans three strands.
First, gold-standard resources are created directly in the target language or via human translation/editing, yielding high in-language validity but slower coverage growth and limited cross-task parallelism.
Examples include ScandEval (Scandinavian) \citep{nielsen_2023}, NorBench (Norwegian) \citep{samuel_norbench_2023}, SuperGLEBer (German) \citep{pfister_2024}, Evalita-LLM (Italian) \citep{Evalita-LLM_2025}, IberoBench (Iberian) \citep{baucells_2025}, as well as BenCzechMark (Czech) \citep{fajcik_2025}, LLMzSz{\L} (Polish) \citep{LLMzSz_2025}, Plutus (Greek finance) \citep{peng_plutus_2025}, and CroissantLLM (French) \citep{faysse_2025}.
Task-specific multilingual benchmarks such as MLQA \citep{lewis-etal-2020-mlqa}, XNLI \citep{conneau-etal-2018-xnli}, XCOPA \citep{ponti_2020_xcopa}, XQuAD \citep{artetxe_2020_cross} and TyDi QA \citep{clark_2020_tydi} offer deep, domain-focused insights but cover a narrow slice of capabilities, which limits cross-task comparability and pan-European parallelism.

Second, machine-translated with human quality assurance (QA) suites broaden language reach while retaining human curation on a subset.
Global-MMLU~\citep{singh_2025_global} provides MT variants (Google Translate) and human-edited subsets for selected languages, enabling reference-based checks.

Third, there are machine-translated with limited or no documented QA resources that emphasize scalability.
EU20~\citep{thellmann_2024} translates five complementary task families -- knowledge-heavy QA (ARC; \citealp{DBLP:journals/corr/abs-1803-05457}), mathematical reasoning (GSM8K; \citealp{cobbe2021gsm8k}), commonsense (HellaSwag; \citealp{DBLP:conf/acl/ZellersHBFC19}), NLI/knowledge probing (MMLU; \citealp{hendrycks2020measuring}), and truthfulness (TruthfulQA; \citealp{lin-etal-2022-truthfulqa}) -- into 20 European languages using DeepL, maximizing parallelism across tasks and languages.
Okapi~\citep{lai-etal-2023-okapi} provides ChatGPT-based translations of ARC, HellaSwag, MMLU, and TruthfulQA into 31 languages, spanning both EU and non-EU languages.
In both cases, a systematic QA pass is not the primary focus of the initial releases.

Against this landscape, our contribution is an automated QA layer for EU20 that is cost-efficient, scalable, and complementary to expert review.
We focus on EU20 because its five task families and 20 languages offer the parallelism required for cross-lingual evaluation at scale.
Our automated QA increases the reliability of this testbed without claiming to replace human-curated gold data.

With respect to other automated translation quality assessment tasks, the WMT translation quality estimation and metrics shared tasks (\citealp{lavie_2025}; \citealp{zerva_2024}; \citealp{freitag_2024}; etc.) consider translations of ``flat'' text passages.
In comparison, LLM evaluation examples have a structure which has to be preserved during translation.
Furthermore, in our work, we also consider consistency across multiple (target) languages as a dataset-level quality criterion.

\section{Structural Diagnostics and Dataset Maintenance for EU20}
\label{sec:struct_audit_repair}

In this section we follow \citet{thellmann_2024}, who introduced EU20 by translating five established English benchmarks into 20 European languages, and provide a structural assessment of the released corpora (Section~\ref{sec:structural_analysis}), applying targeted updates and completions where needed (Section~\ref{sec:correction_completion}).

\subsection{Structural Analysis}
\label{sec:structural_analysis}

We verify the integrity of all translated samples along four criteria:
\begin{enumerate}[A)]
    \item \textbf{Answer-index alignment (for multiple-choice tasks):} Language-independent indices of the correct choice(s) match the English original.
    \item \textbf{Field completeness:} Essential fields are non-empty (``question''; ``choices'' for multiple choice; ``answer'' for generative tasks).
    \item \textbf{Split/subset consistency:} For uniquely identifiable samples, split and subset mirror the English version.
    \item \textbf{Cross-language coverage:} The sample exists across all 20 translations for the evaluated splits.
\end{enumerate}

Criterion A) is satisfied for all datasets.
Table~\ref{tab:dataset_stats_missing} summarizes the remaining criteria:
\(N_C\) is the number of samples violating criterion B), \(N_T\) is the number of samples fulfilling criterion C) across the 20 target languages, and \(N_L\) is the number of samples violating criterion D) after removing samples that violate the other criteria.

Regarding B), we observe missing content primarily in the HellaSwag validation split:
among 10{,}042 English originals, 327 samples (3.26\%) have at least one translation with empty answer options; 257 of these (78\%) affect two or more target languages.
Manual inspection suggests that DeepL can be confused by the context-continuation format used in LM-Eval-Harness when answer options are fragments rather than full sentences, which complicates finding a common prefix across languages.

For criterion C), we did not expect any mismatches, but found two problems that are most likely due to operating errors during the translation:

Firstly, the train splits of our ARC translations, from which the few-shot samples are drawn, consist of a mix of samples from the \texttt{easy} and \texttt{challenge} subsets.
This applies to each of the translated languages, but not to the English version, where the few-shot samples are drawn from the same subset as the sample under test, so comparability of $k$-shot accuracies ($k \geq 1$) between English and non-English languages might be limited.

Secondly, in HellaSwag the per-language \texttt{train} split is not a subset of \texttt{train} but a 99-item subset of \texttt{validation}, which can leak answers into few-shot contexts.
Because the 10 few-shot examples per query are sampled from this small set (\(\approx 100\)) that itself belongs to the validation set (\(\approx 10{,}000\)), the probability that the evaluated item appears in the context is \(\sim 0.1\%\).
The resulting measurement error is proportional to this chance and decreases as the true model accuracy increases.
Even in the worst case (four answer options, true accuracy \(\approx 25\%\)), the overestimation is only about \(0.08\) percentage points (e.g., \(25.00\%\rightarrow 25.08\%\)).
Given this upper bound, we consider the expected fraction of leaks (and thus their impact on reported accuracies) too small to warrant re-evaluating all models on this relatively large and resource-intensive benchmark.

Criterion D) holds for all evaluation splits except HellaSwag and for MMLU-dev.
It is not met for the \texttt{train} splits of ARC, GSM8K, and HellaSwag because sub-splits were selected independently per target language.
Among the HellaSwag validation splits, DE is missing 63 items, and ES, FR and IT 4 items each.

\begin{table}[tbhp]
\tabcolsep=0.28em
\centering
\begin{tabular}{llrrrr}
\textbf{Dataset} & \textbf{Split} & \(\mathbf{N_\text{en}}\) & \(\mathbf{N_T}^{\phantom{\ast\ast}}\)  & \(\mathbf{N_C}\) & \(\mathbf{N_L}\) \\\midrule
ARC              & train          &                    3,370 &   6,420\(^{\ast\phantom{\ast}}\) &                8 &            6,420 \\
ARC              & val            &                      869 &  17,380\(^{\phantom{\ast\ast}}\) &               17 &                0 \\
ARC              & test           &                    3,548 &  70,960\(^{\phantom{\ast\ast}}\) &              109 &                0 \\
GSM8K            & train          &                    7,473 &   2,288\(^{\phantom{\ast\ast}}\) &                0 &            2,288 \\
GSM8K            & test           &                    1,319 &  26,380\(^{\phantom{\ast\ast}}\) &                0 &                0 \\
He.Sw.           & train          &                   39,905 &   1,980\(^{\ast\ast}\)           &               10 &            1,980 \\
He.Sw.           & val            &                   10,042 & 200,765\(^{\phantom{\ast\ast}}\) &            1,039 &            1,185 \\
MMLU             & test           &                   14,042 & 280,840\(^{\phantom{\ast\ast}}\) &              678 &                0 \\
MMLU             & dev            &                      285 &   5,700\(^{\phantom{\ast\ast}}\) &                3 &                0 \\
Tr.QA            & val            &                      817 &  16,340\(^{\phantom{\ast\ast}}\) &                3 &                0 \\
\end{tabular}
\caption{
 \(\mathbf{N_\text{en}}\): \#English samples.
\(\mathbf{N_T}\): \#Non-English samples.
\(\mathbf{N_C}\): \#Non-English samples with missing content.
\(\mathbf{N_L}\): \#Non-English samples with other Non-English version(s) missing.
\(^\ast\)3,060 samples from the respective other subset.
\(^{\ast\ast}\)All samples present translated from validation (val) split.
}
\label{tab:dataset_stats_missing}
\end{table}

\begin{figure*}[t]
  \centering
  \includegraphics[width=0.97\textwidth]{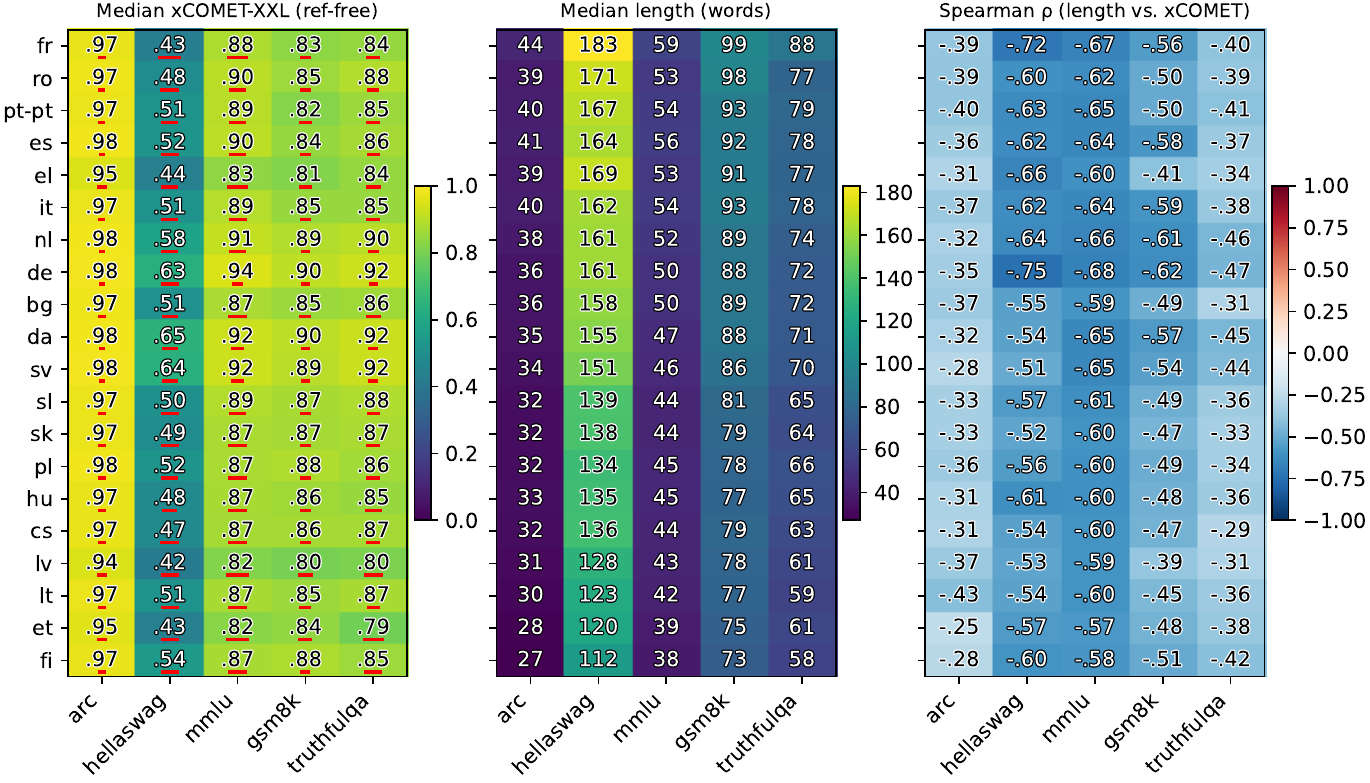}
  \caption{
  EU20 reference-free quality landscape.
  Left: median xCOMET-XXL per language$\times$dataset on a unified $[0,1]$ scale; short in-cell tick encodes IQR ($Q_3{-}Q_1$).
  Middle: median target-side sentence length (words).
  Right: Spearman correlation ($\rho$) between length and score (negative $\rho$ indicates lower scores for longer outputs).
  Rows are aligned across panels and sorted by the language-wise median across datasets.}
  \label{fig:eu20-ref-free-heatmap}
\end{figure*}

\subsection{Corrections and Completion}
\label{sec:correction_completion}

Based on the structural findings, we repair defective entries and complete missing ones via targeted, sample-level re-translation:
\begin{itemize}
    \item \textbf{Scope selection.} We operate either on full split/subset combinations or on JSONL manifests that enumerate individual sample IDs (with associated split/subset and target language) for re-translation. This enables surgical fixes without reprocessing entire corpora.
    \item \textbf{Provenance guarantees.} For each sample, we ensure that the record contains sufficient ID fields so its lineage to the English original is unambiguous before and after modification.
    \item \textbf{Update policy.} Only fields flagged as defective (missing/empty or structurally inconsistent) are overwritten. Intact fields remain unchanged. All edits are logged in an extended diagnostics format alongside the EU20-conform outputs to support auditability and regression checks.
\end{itemize}

We reuse a minimal translation processor primarily for correction/augmentation rather than bulk creation.
The design allows alternative engines, but we currently use DeepL with formatting equivalent to the original setup except for XML escaping.

For each dataset we (i) extract translatable fragments, (ii) ensure/normalize ID/key fields, and (iii) write translated fragments back.
Fragments are serialized into a single XML string (\texttt{Frag\_1<x>SEP</x>} \(\ldots\) \texttt{<x>SEP</x>Frag\_n}) with XML-escaping.
The DeepL API is configured to ignore \texttt{<x>}, and the response is de-serialized to a fragment list.

We support API-level batching and cache formatted inputs/outputs (key: formatted source; value: formatted target) to avoid duplicate calls during iterative fixes.
Outputs are emitted as (a) EU20-conform JSON and (b) an extended diagnostics format for quality control and diffing.

For HellaSwag-style continuations, we optionally reformat options (e.g., prefix repeated context) to reduce empty-field failures, validate completeness post-translation, and route failed/ambiguous cases to manual inspection queues recorded in the diagnostics.

\section{EU20 Translation Quality Profiling with a Neural QE Metric}
\label{sec:eu20-analysis}

In this section, we profile translation quality with xCOMET-XXL, a neural QE metric (reference-free and reference-based), in three steps across tasks and systems:
First, we build a reference-free quality landscape over the EU20 benchmark suite, complemented by a length profile and a length--quality correlation analysis (Section~\ref{subsec:eu20-ref-free}).
Second, we run paired, reference-free comparisons of EU20 versus Okapi across three suites and ten languages, reporting median gaps and win-rates with paired-bootstrap confidence intervals (Section~\ref{subsec:eu20-okapi-ref-free}).
Third, on MMLU we compare EU20, Okapi, and the human-edited Global-MMLU via average-rank testing, and we also perform a reference-based comparison against the human-edited references to validate the trends (Section~\ref{subsec:mmlu-ref-free-3sys} and \ref{subsec:mmlu-ref-based}).

\subsection{Quality Landscape}
\label{subsec:eu20-ref-free}

\paragraph{Method.}
In this comparison, we present a quality landscape of the EU20 translations across five widely used benchmarks: ARC, HellaSwag, MMLU, GSM8K, and TruthfulQA, using the reference-free xCOMET-XXL quality estimator.
To better interpret the results we posit two hypotheses:
(i) longer outputs tend to receive lower xCOMET-XXL scores;
(ii) scores follow a pattern driven by dataset design, with standardized QA (e.g., ARC, MMLU) yielding higher medians than open-ended continuations (HellaSwag).

We test (i) by computing target-side word-count medians for each language-dataset combination, then estimating Spearman's $\rho$ between length and score.
Hypothesis (ii) is treated as a design-based rationale consistent with the observed pattern rather than a tested causal claim.
The results are shown in Figure~\ref{fig:eu20-ref-free-heatmap} (left heatmap) as a $20{\times}5$ matrix (languages $\times$ datasets) on a unified $[0,1]$ scale.
Each cell reports the median score, a short in-cell tick encodes the interquartile range (IQR, $Q_3{-}Q_1$), and rows are sorted by the language-wise median across datasets.
The same figure also presents median sentence-length statistics (middle heatmap) and Spearman correlations between length and score (right heatmap).

\paragraph{Results and discussion.}
Figure~\ref{fig:eu20-ref-free-heatmap} (left) shows the median xCOMET-XXL scores as a $20{\times}5$ matrix. ARC is highest overall (.,97-.,98), HellaSwag is lowest (.,42-.,65), and MMLU (.,83-.,94), GSM8K (.,80-.,90), and TruthfulQA (.,79-.,92) lie in between.
This pattern is consistent with Hypothesis~(ii): standardized QA prompts (ARC, MMLU) tend to yield more semantically aligned translations, whereas open-ended continuations (HellaSwag) induce greater lexical and structural variability.
Across languages, \textsc{de}/\textsc{da}/\textsc{sv} lead (median across datasets $\approx$.,87-.,88), followed by \textsc{nl} (.,85) and a mid-cluster (\textsc{pl}/\textsc{fi}/\textsc{sl}/\textsc{es}/\textsc{ro}, $\approx$.,82); \textsc{el}/\textsc{et} ($\approx .77$) and \textsc{lv} (.,75) trail.
In addition to median levels, longer IQR ticks on HellaSwag and MMLU indicate greater within-dataset variability: HellaSwag combines lower medians with high dispersion (uneven outputs), while MMLU shows higher medians yet similarly wide spread (heterogeneous subjects, formats, and length effects).

Figure~\ref{fig:eu20-ref-free-heatmap} (middle) reports median output lengths, supporting Hypothesis~(i)'s premise about verbosity: HellaSwag is longest (medians $\sim$112-183 words), followed by GSM8K (73-99), TruthfulQA (58-88), and MMLU (38-59); ARC is shortest (27-44).
This aligns with task design, where open-ended reasoning naturally produce longer outputs than fixed-format multiple-choice questions.

Figure~\ref{fig:eu20-ref-free-heatmap} (right) shows Spearman correlations between length and xCOMET-XXL.
Correlations are predominantly negative, indicating that longer translations tend to receive lower quality estimates.
Effect sizes vary by dataset: ARC weak-moderate ($\rho\approx-0.25$ to $-0.43$), TruthfulQA moderate ($\rho\approx-0.29$ to $-0.47$), GSM8K and MMLU stronger ($\rho\approx-0.39$ to $-0.68$), and HellaSwag strongest (often $<-0.60$; e.g., \texttt{de} $\rho\approx-0.75$), with $p \ll .001$ in most cells.
Overall, the dataset pattern remains the dominant signal, while length effects help explain within-cell dispersion and some cross-dataset differences.

\subsection{EU20 vs.~Okapi}
\label{subsec:eu20-okapi-ref-free}

\paragraph{Method.}
To directly compare the quality of our translations against Okapi across tasks, we compute reference-free xCOMET-XXL scores on the shared subset of items for three representative benchmarks ARC, HellaSwag, and MMLU, and ten overlapping languages (CORE-10: \textsc{da, de, es, fr, hu, it, nl, ro, sk, sv}).
Working on the paired item overlap (identical $(\textit{language, subset, split, id})$) ensures a fair comparison, as both translation systems are evaluated on exactly the same segments.
For each (language, dataset) pair, we compute the median score difference $\Delta = \mathrm{median}(EU20) - \mathrm{median}(Okapi)$, where positive $\Delta$ indicates higher predicted adequacy/fluency for EU20.

To visualize these results, we use a heatmap with a diverging, zero-centered color scale fixed to a symmetric range, enabling comparison across datasets and languages.
Each cell also reports the win-rate $P(EU20 > Okapi)$, i.e., the proportion of samples for which EU20 receives a higher xCOMET-XXL score than Okapi.
This provides a complementary, distribution-sensitive indicator of relative system quality beyond the median.

\begin{figure}[t]
    \centering
    \includegraphics[width=0.85\columnwidth]{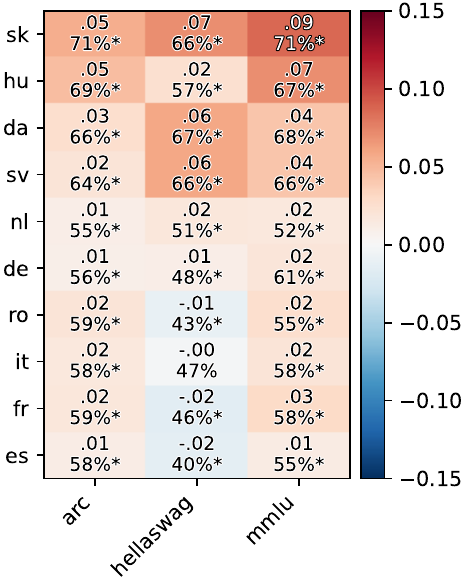}
    \caption{EU20 vs. OKAPI xCOMET-XXL reference-free quality comparison per language$\times$dataset. Cells report the median difference $\Delta=\mathrm{median}(EU20)-\mathrm{median}(Okapi)$ on the paired overlap and the win-rate (\% items where EU20 $>$ Okapi). Positive $\Delta$ favors EU20.}
  \label{fig:eu20-okapi-delta-heatmap}
\end{figure}

We assess the statistical reliability of $\Delta$ using a paired bootstrap, a standard distribution-free method in MT/NLP for system-level score comparisons \citep{Koehn2004,Dror2018,EfronTibshirani1994}.
We resample paired segment scores $B \approx 5000$ times, recompute the median difference for each replicate, and form a $(1-\alpha)$ confidence interval from the empirical quantiles of the bootstrap distribution.
Figure~\ref{fig:eu20-okapi-delta-heatmap} summarizes the paired, reference-free comparison of EU20 vs.~Okapi on ARC/HellaSwag/MMLU across the CORE-10 languages.
Given the large per-cell sample sizes (e.g., $\sim$1.4k for ARC, $\sim$7.3k for HellaSwag, $\sim$11.4k for MMLU), the intervals are typically narrow and differences often reach statistical significance.
A star marks cells where the 95\% CI does not include zero, indicating a statistically significant difference.

\paragraph{Results and discussion.}
On ARC and MMLU, EU20 is ahead in every language: the median score advantage is typically $\Delta\!\approx\!.009$-$.054$ on ARC and $.012$-$.086$ on MMLU.
These gains are small in absolute terms but statistically reliable: bootstrap confidence intervals for ($\Delta$) do not include $0$, and win-rates are mostly $[0.55, 0.71]$ with intervals above $0.5$ (i.e., EU20 wins on a clear majority of segments).
Because scores are bounded in $[0,1]$, differences between strong systems are compressed, so improvements of a few hundredths (together with win-rates $>0.5$ and CIs excluding $0$) are therefore meaningful when consistent.

HellaSwag is mixed: EU20 leads in \textsc{sk}/\textsc{sv}/\textsc{da}/\textsc{hu}/\textsc{nl}/\textsc{de}, while Okapi (Google Translate) edges out in \textsc{es}/\textsc{fr}/\textsc{ro}; \textsc{it} is at parity (small, non-significant $\Delta$).
One plausible interpretation is a better stylistic fit of Okapi on colloquial continuations in some Romance languages.
We treat this as an observation consistent with the data rather than a causal claim.

Two signals support the overall conclusion that in the majority of language-task combinations, the EU20 translations have higher xCOMET-XXL scores on average than the Okapi translations:
(i) win-rates ($>0.5$) show that EU20 wins on a majority of segments, and
(ii) large per-cell sample sizes ($\sim 1.4k$ on ARC, $\sim 7.3k$ on HellaSwag, $\sim 11.4k$ on MMLU) yield narrow bootstrap intervals, so many gaps are statistically significant.

Finally, this pattern is consistent with the reference-based MMLU analysis (Section~\ref{subsec:mmlu-ref-based}): the ref-free advantage of EU20 over Okapi persists when evaluated against human-edited references and is significant in 4/5 languages.
This triangulation suggests the effect is systematic rather than specific to a single evaluation mode.

\subsection{MMLU Ranks: EU20, Okapi, Global}
\label{subsec:mmlu-ref-free-3sys}

\paragraph{Method.}
We compare three translation sources EU20 (DeepL), Okapi (ChatGPT), and Global-MMLU (Google Translate, human-edited) on MMLU in five languages \{\textsc{de, es, fr, it, ro}\}, using the triple item overlap (segments with identical language, subset, split, and id).
On this shared set of items, we compute per-system median xCOMET-XXL scores, then convert the three medians to ranks (1\,=\,highest median; ties share average rank).

\begin{figure}[t]
  \centering
  \includegraphics[width=0.98\columnwidth]{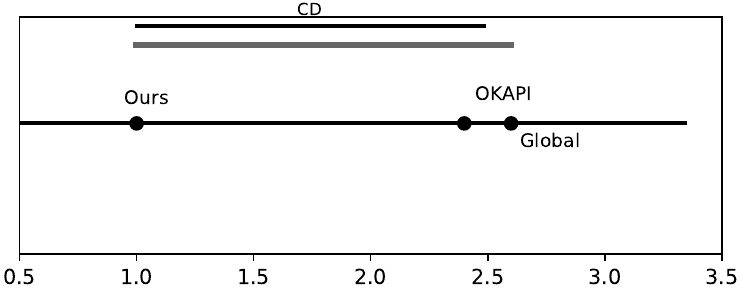}
  \caption{Critical-difference (CD) diagram on MMLU (ref-free). Points are systems' average ranks across five languages (lower is better). Thin bars show Nemenyi intervals (avg$\pm$CD/2; $\alpha{=}0.05$, $k{=}3$, $N{=}5$). A grey bridge links systems that are not significantly different (no bridge = significant).}
  \label{fig:mmlu-cd-plot}
\end{figure}

To determine whether observed rank gaps are more than descriptive, we use the standard ML/NLP workflow -- Friedman's omnibus test (blocks\,=\,languages) followed by the Nemenyi all-pairs test on mean ranks, to determine which systems differ across datasets \citep{Demsar2006,Garcia2008}.
The resulting critical difference (CD) is the minimum gap between average ranks that must be exceeded for a pair to be considered significantly different at $\alpha{=}0.05$ \citep{Demsar2006}.
Graphically, the CD is shown as Nemenyi intervals (avg$\pm$CD/2) where overlap indicates ``not significantly different'' and non-overlap indicates a significant difference.
Figure~\ref{fig:mmlu-cd-plot} visualizes average ranks with Nemenyi intervals and bridges indicating pairs that are not significantly different.

\paragraph{Results and discussion.}
On the triple overlap (2{,}378 items per language), our translation attains rank~1 in all five languages (average rank\,=\,1.00), while Okapi and Global-MMLU occupy ranks 2-3 (see Table~\ref{tab:mmlu-ref-free-medians-ranks} for medians and ranks).
With $k{=}3$ systems and $N{=}5$ languages, the Nemenyi CD is $\approx1.48$.
The EU20-Global average-rank gap is 1.60\,{>}~CD~$\Rightarrow$ significant, whereas EU20-OKAPI (1.40\,{\(\le\)}\,CD) and OKAPI-Global (0.20\,{\(\le\)}\,CD) are not significant (see Figure~\ref{fig:mmlu-cd-plot}).
The per-language medians in Table~\ref{tab:mmlu-ref-free-medians-ranks} show the same ordering.
We hypothesize that minor differences in paraphrasing/verbosity and morphology handling on EN$\to$\{de, es, fr, it, ro\} make EU20/DeepL align slightly better with COMET's adequacy/fluency signals.
This finding is consistent with the reference-based results (see Section~\ref{subsec:mmlu-ref-based}).

\begin{table}[t]
  \centering
  \footnotesize
  \tabcolsep=0.20em
  \begin{tabular}{lrrrrrrr}
  \toprule
lang & $m_{\mathrm{EU20}}$ & $m_{\mathrm{Okapi}}$ & $m_{\mathrm{Global}}$ & items & $r_{\mathrm{EU20}}$ & $r_{\mathrm{Okapi}}$ & $r_{\mathrm{Global}}$ \\
  \midrule
  de &	.96	& .94	& .95	& 2378	& 1	& 3	& 2 \\
  es &	.93	& .92	& .91	& 2378	& 1	& 2	& 3 \\
  fr &	.91	& .89	& .88	& 2378	& 1	& 2	& 3 \\
  it &	.92	& .90	& .91	& 2378	& 1	& 3	& 2 \\
  ro &	.93	& .91	& .88	& 2378	& 1	& 2	& 3 \\
  \bottomrule
  \end{tabular}
  \caption{Per-language medians and ranks on the triple overlap (MMLU, ref-free).
  For each language, we report median xCOMET-XXL for EU20, Okapi, and Global-MMLU, the number of common items, and per-language ranks (1\,=\,best).}
  \label{tab:mmlu-ref-free-medians-ranks}
\end{table}

\subsection{Ref-Based MMLU: EU20 vs.~Okapi}
\label{subsec:mmlu-ref-based}

\begin{table}[t]
  \centering
  \footnotesize
  \tabcolsep=0.30em
  \renewcommand{\arraystretch}{0.95}
  \begin{tabular}{llllll}
    \toprule
    lang & $m_{\mathrm{EU20}}^{\mathrm{ref}}$ & $m_{\mathrm{Okapi}}^{\mathrm{ref}}$ & $\Delta_{\mathrm{ref}}$ &  CI[low,high] &  items \\
    \midrule
    it & .92 & .89 & .029 & [.0195, .0398] & 2342 \\
    de & .96 & .93 & .026 & [.0218, .0320] & 2342 \\
    fr & .90 & .88 & .025 & [.0150, .0344] & 2342 \\
    ro & .92 & .91 & .015 & [.0070, .0232] & 2342 \\
    es & .91 & .91 & -.004 & [-.0237, .0158] &  516 \\
    \bottomrule
  \end{tabular}
  \caption{Per-language reference-based medians and deltas on MMLU.
  For each language, we report $\mathrm{median}(EU20_{\text{ref}})$, $\mathrm{median}(Okapi_{\text{ref}})$, $\Delta_{\text{ref}}$, 95\% CI, and $n$ of common items.}
  \label{tab:mmlu-ref-based-deltas}
\end{table}

\begin{figure}[t]
  \centering
  \includegraphics[width=0.8\columnwidth]{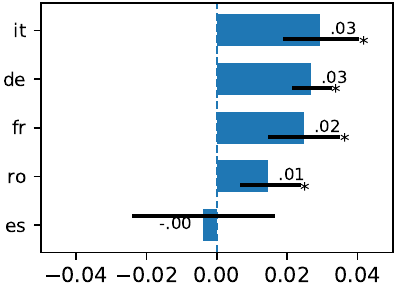}
  \caption{$\Delta_{\text{ref}}$ (EU20$-$Okapi) of reference-based xCOMET-XXL on MMLU (reference = Global-MMLU).
  Bars show $\Delta_{\text{ref}}$ per language with 95\% paired bootstrap CIs.
  Zero line indicates parity.
  Sorted by $\Delta_{\text{ref}}$.
  Common items only.}
  \label{fig:mmlu-ref-based-delta}
\end{figure}

\paragraph{Method.}
We compare two MMLU translation variants in a reference-based setting: EU20 (DeepL-derived) vs.~Okapi (ChatGPT-derived), using Global-MMLU as the human-edited reference.
To ensure strict comparability, we restrict both systems to the paired item overlap, i.e., the common set of segments with identical language, subset, split and id, present in both sources, for the five languages with vetted references \{\textsc{de, es, fr, it, ro}\}.
On this shared set, we summarize the effect per language as \(\Delta_{\mathrm{ref}} \;=\; \mathrm{median}\!\big(EU20_{\mathrm{ref}}\big)\;-\;\mathrm{median}\!\big(Okapi_{\mathrm{ref}}\big),\) where larger values favor EU20.
Uncertainty is quantified with a 95\% paired bootstrap CI on $\Delta_{\mathrm{ref}}$ (resampling the same indices in both systems; $B\!\approx\!5000$) to avoid distributional assumptions \citep{EfronTibshirani1994}.
Figure~\ref{fig:mmlu-ref-based-delta} displays horizontal bars per language with CIs and a vertical zero line indicating parity ($\Delta_{\mathrm{ref}}{=}0$), using fixed axes for visual comparability and sorting languages by $\Delta_{\mathrm{ref}}$.
The corresponding per-language medians, deltas, CIs, and item counts are reported in Table~\ref{tab:mmlu-ref-based-deltas}.

\paragraph{Results and discussion.}
Across the five languages, EU20 is significantly better than Okapi in 4/5 cases (see Figure~\ref{fig:mmlu-ref-based-delta} and Table~\ref{tab:mmlu-ref-based-deltas}):
\textsc{it} ($\Delta_{\mathrm{ref}}{=}.029$, 95\% CI $[.020-.040]$),
\textsc{de} ($.027 [.022-.032]$),
\textsc{fr} ($.025 [.015-.034]$),
\textsc{ro} ($.015 [.007-.023]$)
\textsc{es} shows no significant difference ($\Delta_{\mathrm{ref}}{=}{-}.004$, CI [${-}.024$, .016]).
Effect sizes are small-to-moderate (hundredths) but consistent over large item counts ($n$).

The wider CI in \textsc{es} reflects the much smaller paired overlap ($n{=}516$ vs.~2{,}342 elsewhere).
These findings complement the ref-free comparison in Section~\ref{subsec:mmlu-ref-free-3sys}: although EU20 differs significantly from Global-MMLU in the reference-free setting (reflecting stylistic or phrasing divergences that xCOMET-XXL judges more favorably), it still aligns more closely with the human-edited Global-MMLU translations than Okapi does when those translations are used as explicit references.
In other words, EU20 is not identical to Global-MMLU, but it captures key aspects of phrasing and morphology that bring it significantly closer to the reference than Okapi across four of five languages.

\section{Translation Error Landscape for EU20 from Span-Level Judgments}
\label{subsec:eu20_gemba}

As a complement to the sentence-level quality profiling with xCOMET-XXL -- a neural, dual-mode QE metric (reference-free/-based; Section~\ref{sec:eu20-analysis}) -- we conduct a span-level error profiling of EU20 translations.
We adopt an LLM-as-a-judge TQE setup based on GEMBA-ESA~\citep{kocmi_2023,kocmi_2024} and the MQM taxonomy~\citep{lommel_2013} to annotate translation errors with category and severity.
The two perspectives are complementary: xCOMET-XXL provides scalar sentence-level quality signals, whereas GEMBA-ESA exposes an interpretable error structure (what went wrong, and how severe) at span level.

\paragraph{Method.}
We adapt GEMBA-ESA with a structured JSON output and multilingual few-shot prompts that instruct the model to detect span-level errors and label them with MQM categories and a severity (major if meaning is impaired, minor if generally understandable).
We use three independent LLM annotators -- GPT-4o-mini\footnote{\href{https://platform.openai.com/docs/models/gpt-4o-mini}{platform.openai.com/docs/models/gpt-4o-mini}}, Llama-4 Scout\footnote{HF: \href{https://huggingface.co/meta-llama/Llama-4-Scout-17B-16E-Instruct}{meta-llama/Llama-4-Scout-17B-16E-Instruct}}, and Mistral-Large-Instruct-2411\footnote{HF: \href{https://huggingface.co/mistralai/Mistral-Large-Instruct-2411}{mistralai/Mistral-Large-Instruct-2411}}  -- all prompted with our adapted GEMBA-ESA prompt.
For comparability across languages and tasks, we aggregate the MQM span labels produced by the LLM annotators into the high-level categories:
\texttt{A+M} (accuracy+mistranslation), \texttt{F} (fluency/style), and \texttt{O} (other).

\begin{figure}[t]
  \centering
  \includegraphics[width=0.98\columnwidth]{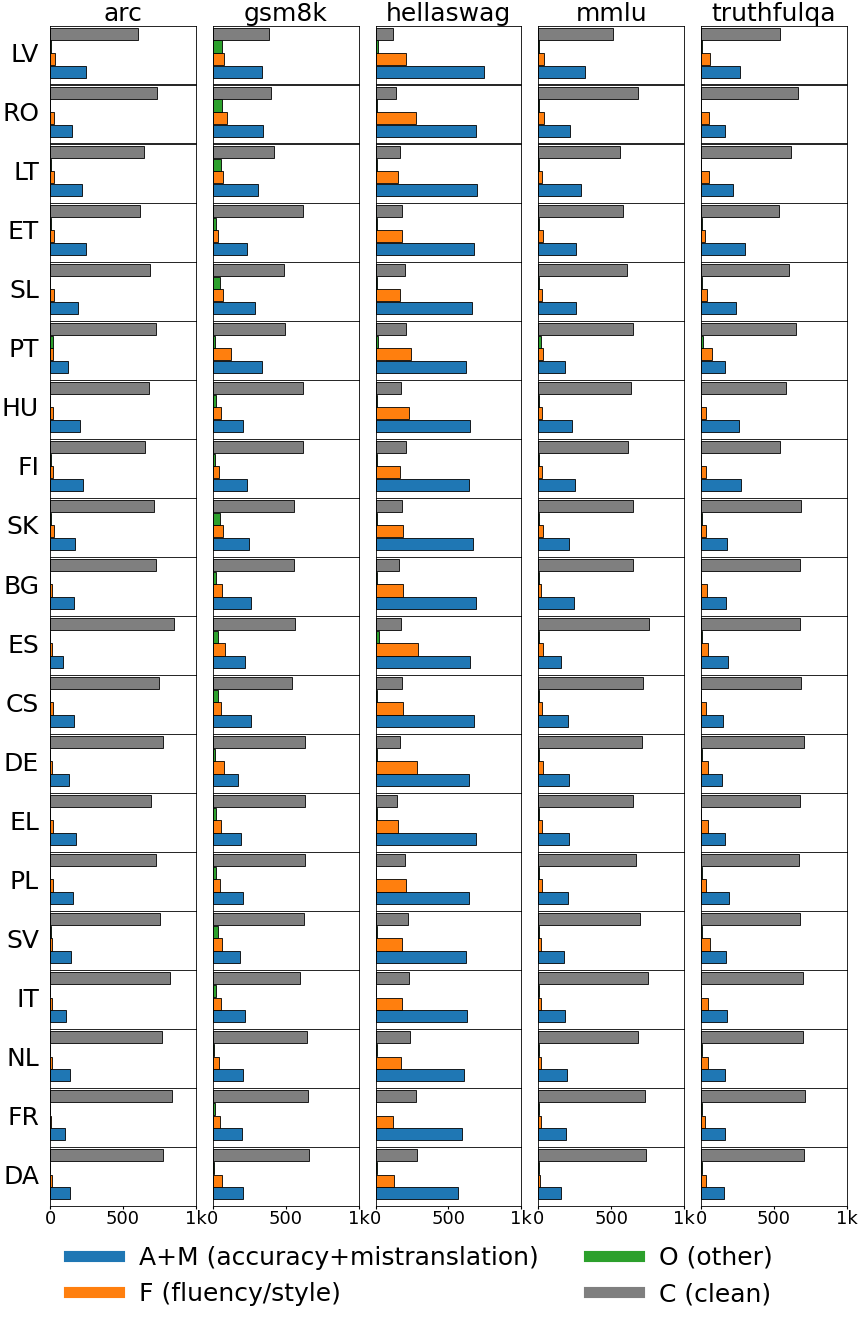}
   \caption{
   EU20 error overview per \(\text{language}\times\text{dataset}\).
   Each cell shows four horizontal bars: \texttt{A+M}, \texttt{F}, \texttt{O}, and \texttt{Clean}.
   Error rates per 1,000 items.
   }
  \label{fig:eu20-overview}
\end{figure}

For each item (one translated benchmark entry) and category (\texttt{A+M}, \texttt{F}, \texttt{O}), we set \(\mathrm{maj}(i,T,S)=1\) iff at least two annotators flagged any span of type \(T\) with severity \(S\) (\(S\in\{\text{minor},\text{major}\}\); if both severities would apply, we keep major to avoid double counting).
For Figure~\ref{fig:eu20-overview}, we collapse severities and mark a category as present if either minor or major reached majority.
\texttt{CLEAN} holds if a majority judged ``no error'' (items without error spans count as no error for that annotator).
We report rates per 1{,}000 items for each category and \texttt{CLEAN} for every language and dataset.

Although Figure~\ref{fig:eu20-overview} collapses severities for space, Table~\ref{tab:eu20-severity-by-ds} reports the share of major/minor within each category per dataset.
We pool counts across all 20 languages: for each dataset and category, the major percentage is the share of agreed major errors out of all agreed errors in that category (major+minor), and minor percentage is the complementary share.

\paragraph{Results and discussion.}
A consistent pattern emerges across languages: (i) HellaSwag shows the highest \texttt{A+M} rates (e.g., \texttt{LV}: 744/1k, \texttt{RO}: 691/1k, \texttt{BG}: 693/1k), with \texttt{F} smaller (typically \(\sim\)120-290/1k) and \texttt{O} low (single digits to \(\sim\)20/1k).
Clean is correspondingly low (often \(\sim\)120-285/1k).
(ii) GSM8K and MMLU sit mid-range in \texttt{A+M} (\(\sim\)170-347/1k and \(\sim\)150-322/1k, respectively) with \texttt{CLEAN} mostly \(\sim\)500-750/1k.
(iii) ARC is comparatively clean (\texttt{A+M} \(\sim\)90-250/1k; \texttt{CLEAN} \(\sim\)700-850/1k).
(iv) TruthfulQA shows moderate \texttt{A+M} (146-303/1k) and high \texttt{CLEAN} (535 - 709/1k).
\texttt{CLEAN} is generally higher in Germanic/Romance languages (e.g., \texttt{DA}/\texttt{NL}/\texttt{SV}/\texttt{DE}/\texttt{FR}) than in Baltic/Balkan languages.
However, these cross-language gaps are modest compared to the dataset effect -- differences between datasets (e.g.~HellaSwag vs.~ARC) are substantially larger than differences between languages within a dataset.
Overall, \texttt{A+M} dominates the error mass, \texttt{F} is secondary, and \texttt{O} marginal.

As shown in Table~\ref{tab:eu20-severity-by-ds}, \texttt{A+M} errors are predominantly major across datasets -- highest on HellaSwag (87.6\% maj), followed by GSM8K (83.4\%) and MMLU (81.9\%).
ARC and TruthfulQA are lower yet still predominantly major (73.9\% and 75.4\%).
By contrast, \texttt{F} and \texttt{O} are mostly minor in all datasets (e.g., HellaSwag  \texttt{F} 2.0/98.0, \texttt{O} 4.1/95.9).
\texttt{CLEAN} is lowest on HellaSwag (19.3\%) and highest on ARC (72.6\%)

\begin{table}[t]
  \centering
  \footnotesize
  \tabcolsep=0.35em
  \renewcommand{\arraystretch}{1.05}
  \begin{tabular}{lcccc}
    \toprule
      & \texttt{AM} & \texttt{F} & \texttt{O} & \texttt{CLEAN} \\
                     & \texttt{maj/min}        & \texttt{maj/min}        & \texttt{maj/min}       & \texttt{-/-}\\
    \midrule
    arc        & 73.9/26.1 & 3.7/96.3  & 0.9/99.1  & 72.6 \\
    gsm8k      & 83.4/16.6 & 6.1/93.9  & 2.2/97.8  & 56.7 \\
    hellaswag  & 87.6/12.4 & 2.0/98.0  & 4.1/95.9  & 19.3 \\
    mmlu       & 81.9/18.1 & 5.1/94.9  & 9.4/90.6  & 66.3 \\
    truthfulqa & 75.4/24.6 & 7.3/92.7  & 1.4/98.6  & 65.2 \\
    \bottomrule
  \end{tabular}
  \caption{Share of major/minor (\texttt{maj}/\texttt{min}) severities (in \%) among majority-agreed errors per category (\texttt{A+M}  \texttt{F}, \texttt{O}, \texttt{CLEAN})}
  \label{tab:eu20-severity-by-ds}
\end{table}

The translation error profile/landscape from span-level judgements aligns closely with the sentence-level xCOMET-XXL quality landscape (Section~\ref{sec:eu20-analysis}):
datasets with lower xCOMET-XXL medians (notably HellaSwag) are precisely those with high \texttt{A+M} rates here.
This supports our earlier hypotheses: task design drives difficulty (open-ended continuations induce adequacy/mistranslation pressure), and length effects exacerbate it, producing lower sentence-level quality scores and more adequacy-span agreements.
Conversely, ARC (short, standardized QA) shows high \texttt{CLEAN} and low \texttt{A+M}, matching its high xCOMET-XXL medians.
Taken together, xCOMET-XXL provides scalar, task-sensitive sentence-level quality signals, while GEMBA-ESA reveals where the quality is lost: predominantly accuracy/mistranslation (often major), rather than fluency or style issues.
This suggests that targeted clean-up of high-\texttt{A+M} clusters (especially on HellaSwag) would yield the largest quality gains for EU20.

\section{Conclusion}
We examined the reliability of machine-translated benchmark evaluation at pan-European scale by diagnosing translation quality in EU20.
Our automated QA stack combines (i) a corpus audit and targeted repairs, (ii) COMET-based sentence-level profiling (reference-free and reference-based) with comparisons across translation services, and (iii) span-level LLM-as-a-judge (MQM).
Results converge: HellaSwag has the highest accuracy/mistranslation error mass and lowest COMET; ARC is cleanest; longer outputs correlate with lower quality.
On MMLU, reference-based COMET against human-edited Global-MMLU samples supports ref-free rankings.
Our take-away is pragmatic: automated QA provides quantifiable, diagnostic evidence that helps decide where scarce human QA is most valuable, strengthening cross-lingual comparisons at scale.

\section{Limitations}
Our study is centered on EU20 (five tasks and 20 European languages) so conclusions may not transfer to other domains or non-European languages.
We rely on automated quality assurance as a proxy for human review: COMET and LLM-as-a-judge can introduce metric and judge biases, and span-level MQM may suffer from boundary ambiguity, over-fragmentation of single errors, and unstable severity calibration under prompt/model changes.
Human verification is limited to a reference-based check on a subset of MMLU (human-edited Global-MMLU).
We do not provide large-scale human adjudication across all tasks and languages (out of scope).

Stability remains a concern: results can vary with prompts, seeds, and evolving MT/LLM/COMET versions despite reporting confidence intervals.
Comparisons across translation services (DeepL, ChatGPT, Google) are observational and tied to specific pipelines and time windows.
Small deltas should be interpreted cautiously rather than causally.
Finally, structural audits ensure format integrity (fields, splits, coverage), and TQE-based evaluations cannot guarantee full semantic faithfulness or cultural appropriateness of translations.

\section{Ethical \& Broader Impact}
The ability to evaluate large language models (LLMs) across a wide range of European languages, particularly underrepresented ones, is a critical step toward enhancing inclusivity and accessibility in natural language processing (NLP).
By ensuring that LLMs can perform well in languages beyond English or other high-resource languages, we contribute to a more equitable digital landscape where speakers of less widely spoken languages have equal access to advanced language technologies.
However, this inclusivity brings unique challenges, particularly in achieving benchmarks that are comparable across diverse linguistic and cultural contexts.

\section*{Acknowledgments}
This work was funded by the German Federal Ministry for Economic Affairs and Climate Action (BMWK) through the project OpenGPT-X (project no. 68GX21007D).
The authors acknowledge the financial support by the Federal Ministry of Research, Technology and Space of Germany (BMFTR) and by Sächsische Staatsministerium für Wissenschaft, Kultur und Tourismus in the programme Center of Excellence for AI-research „Center for Scalable Data Analytics and Artificial Intelligence Dresden/Leipzig“, project identification number: ScaDS.AI, and by the BMFTR under grant number 01IS24077A.
The authors gratefully acknowledge the Gauss Centre for Supercomputing e.V. (www.gauss-centre.eu) for funding this project by providing computing time on the GCS Supercomputer JUWELS at Jülich Supercomputing Centre (JSC) as well as the Center for Information Services and High Performance Computing [Zentrum für Informationsdienste und Hochleistungsrechnen (ZIH)] at TU Dresden for providing its facilities for automatic evaluation computations.

\section{Bibliographical References}\label{sec:reference}

\bibliographystyle{lrec2026-natbib}
\bibliography{literature}

\end{document}